# Application of the Second-Order Statistics for Estimation of the Pure Spectra of Individual Components from the Visible Hyperspectral Images of Their Mixture

Sung-Ho Jong   Yong-U Ri   Kye-Ryong Sin[*]

(Department of Chemistry, **Kim Il Sung** University, Daesong district, Pyongyang, DPR Korea)

*E-mail: ryongnam9@yahoo.com

**Abstract:** The second-order statistics (SOS) can be applied in estimation of the pure spectra of chemical components from the spectrum of their mixture, when SOS seems to be good at estimation of spectral patterns, but their peak directions are opposite in some cases. In this paper, one method for judgment of the peak direction of the pure spectra was proposed, where the base line of the pure spectra was drawn by using their histograms and the peak directions were chosen so as to make all of the pure spectra located upwards over the base line. Results of the SOS analysis on the visible hyperspectral images of the mixture composed of two or three chemical components showed that the present method offered the reasonable shape and direction of the pure spectra of its components.

## 1. Introduction

During the past decade the hyperspectral image analysis has emerged as a powerful tool in quality control and identification of various products including foodstuff and in non-destructive study of biomaterials, because it has some superiorities over the traditional methods of chemical analysis in investigation of the composition, spatial distribution and dynamic feature of the chemical components in the samples.

Hyperspectral image analysis gets the information of the sample's chemical constitution and its spatial distribution by analyzing spectral data set(hyperspectral cube) consisted of measured spectral data of every pixels in 2D measuring area.

The analysis of the hyperspectral cube is effective when every pixels have different spectral information.

Meanwhile spectral patterns and intensities corresponding to every pixels in hyperspectral cube depend on the constitution and concentration of the chemical components at the pixels. Consequently, spectral image analysis is applicable to the analysis of the samples containing

heterogeneous distribution of the chemical components

2D concentration distribution of the individual components can be readily determined by the hyperspectral image resolution when FT-IR or Raman spectra of the pure components were used as primary data because these spectra can be clearly distinguished in the measured wavelength intervals. But it is not applicable in the visible - short wave infrared spectra because in this region ($\lambda$=400~1000nm) the spectra of the components can not be easily separated, and consequently the more advanced methods of spectral image analysis are required.

The key in hyperspectral image(in this work we call the spectra of the mixture) analysis is the correct estimation of the number and pattern of the signals from the individual components.[1-5] Of the earlier known methods such as PCA (Principle Component Analysis),[14] the end-member[15], MCR-ALS (Multivariate Curve Resolution - Alternating Least Squares),[16] ICA (Independent Component Analysis)[10], and SOS (Second-Order Statistics)[6-9] for estimation of source signals ICA and SOS are regarded to be fit up to date.

ICA is the best for high accuracy of source signal detection, but the source signals are not supposed to have Gaussian distribution and furthermore difficulty in optimal selection of the objective function limits its application.

SOS is not as accurate as ICA but it is relatively simple because optimal selection of the objective function is not required for the source signal separation. Specifically, it is as accurate as ICA when the source signals have time series structure. Considering the fact that spectral intensities of one chemical component in the two different wavelengths vary in a certain relation, SOS is supposed to be more applicable for the spectra image analysis of the mixtures.

There are some reports on application of SOS for obtaining the pure signal of the individual sources from their mixed ones in mechanical vibrations, acoustics and electric communications,[12,13] but it is unknown on SOS applications for the spectral image analysis of the chemical mixtures.

In this paper presented was one method of choosing the correct peak direction of the pure spectra of the chemical components obtained by SOS analysis of the visible spectra of their mixture.

## 2. Judgment of the correct peak direction of pure spectra of the chemical components

Source signal detection by SOS is based on the singular value decomposition of the time delayed covariance matrix to separate unknown source signals from the space-timely mixed signal.[17]

The equation between the measured signals and the source signals can be written as

$$x(t) = A\, s(t) + N(t) \qquad (1)$$

where t is discrete measuring variable, A is m×n mixing matrix, and

$x(t) = (x_1(t), x_2(t), \cdots, x_m(t))^T$ : vector of m output observations

$$s(t) = (s_1(t), s_2(t), \cdots, s_n(t))^T \quad : \text{ vector of n source signals}$$

$$N(t) = (N_1(t), N_2(t), \cdots, N_m(t))^T : \text{ vector of m measuring noises}$$

AMUSE algorithm for source signal detection by SOS is as follows:

i) to estimate covariance matrix $\hat{R}_x(0)$ of zero-mean output signals,

$$\hat{R}_x(0) = 1/N \sum_{k=1}^{N} x(k) x^T(k) \tag{2}$$

where N is a number of measuring variables.

ii) to calculate eigenvalue decomposition (EVD) or singular value decomposition (SVD) of $\hat{R}_x(0)$,

$$\hat{R}_x(0) = U_x \sum\nolimits_x V_x^T = V_x \Lambda_x V_x^T = V_s \Lambda_s V_s^T + V_N \Lambda_N V_N^T \tag{3}$$

where $V_s = [v_1, v_2, \cdots v_n]$ is a m×n matrix corresponding to ascendingly arranged n dominant eigenvalues $\Lambda_s = \text{diag}\{\lambda_1 \geq \lambda_2 \geq \cdots \lambda_n\}$ and $V_N$ is a m×(m-n) matrix which contains m-n noise eigenvectors corresponding to noise eigenvalue $\Lambda_N = \text{diag}\{\lambda_{n+1} \geq \lambda_{n+2} \geq \cdots \lambda_m\}$.

iii) to apply preliminary whitening conversion,

$$\bar{x}(k) = \Lambda_s^{-\frac{1}{2}} \cdot V_s^T \cdot x(k) = Qx(k) \tag{4}$$

iv) to estimate covariance matrix of vector $\bar{x}(k)$ for special time delay $\tau \neq 0$ and apply SVD for estimated covariance matrix.

$$\hat{R}_{\bar{x}}(\tau) = 1/N \sum\nolimits_{k=1}^{N} \bar{x}(k) \bar{x}^T(k-\tau) = U_{\bar{x}} \cdot \sum\nolimits_{\bar{x}} \cdot V_{\bar{x}}^T \tag{5}$$

v) To estimate source signal and mixing matrix,

$$\hat{A} = Q^+ U_{\bar{x}} = V_s \cdot \Lambda_s^{\frac{1}{2}} \cdot U_{\bar{x}} \tag{6}$$

$$y(k) = \hat{s}(k) = U_{\bar{x}}^T \bar{x}(t) \tag{7}$$

To apply the above SOS analysis procedures in the estimation of the pure spectra of the individual chemical components from the spectra of their mixture, here, proposed was one simple model solution that has two chemical components (P1 and P2) as solutes (for example, an aqueous solution of potassium permanganate $KMnO_4$ and potassium dichromate $K_2Cr_2O_7$) and supposed that the spectra of both of $P_1$ solution and $P_2$ solution were already known as shown in Figure 1.

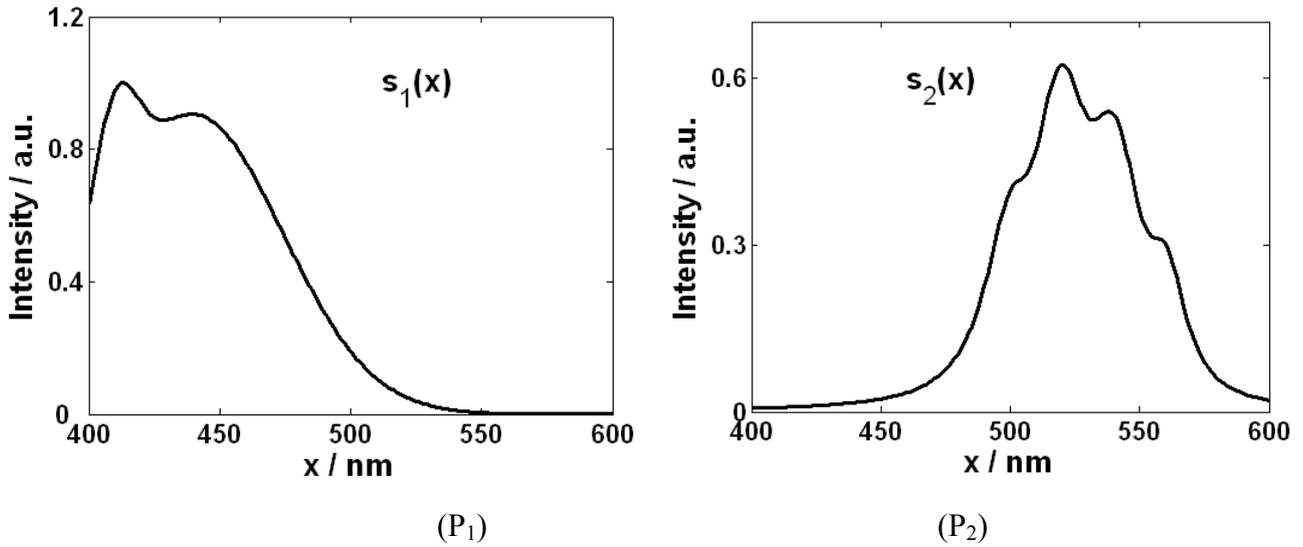

(P₁) (P₂)

**Figure 1.** The initial pure spectra of two components ($P_1$ and $P_2$)

From here, SOS analysis on the above model was carried out by using MATLAB.[18]

It was assumed that the measuring object is consisted of 7 pixels and the mixing matrix A is composed as

$$A = [\,0.2\ \ 0.8;\ 0.8\ \ 0.2;\ 0.4\ \ 0.6;\ 0.3\ \ 0.7;\ 0.9\ \ 0.1;\ 0\ \ 1;\ 1\ \ 0\,]$$

and the measured spectrum of the $P_1$-$P_2$ mixture is modeled by

$$y(x) = A\,s(x),\quad s(x) = (s_1(x), s_2(x))^T$$

In this case, the estimated spectra of the $P_1$-$P_2$ mixture and its character were shown in Figure 2, where 2-a is the change of y(x) according to x and 2-b is the change of column vectors of A according to pixels, showing the concentration distribution of the components.

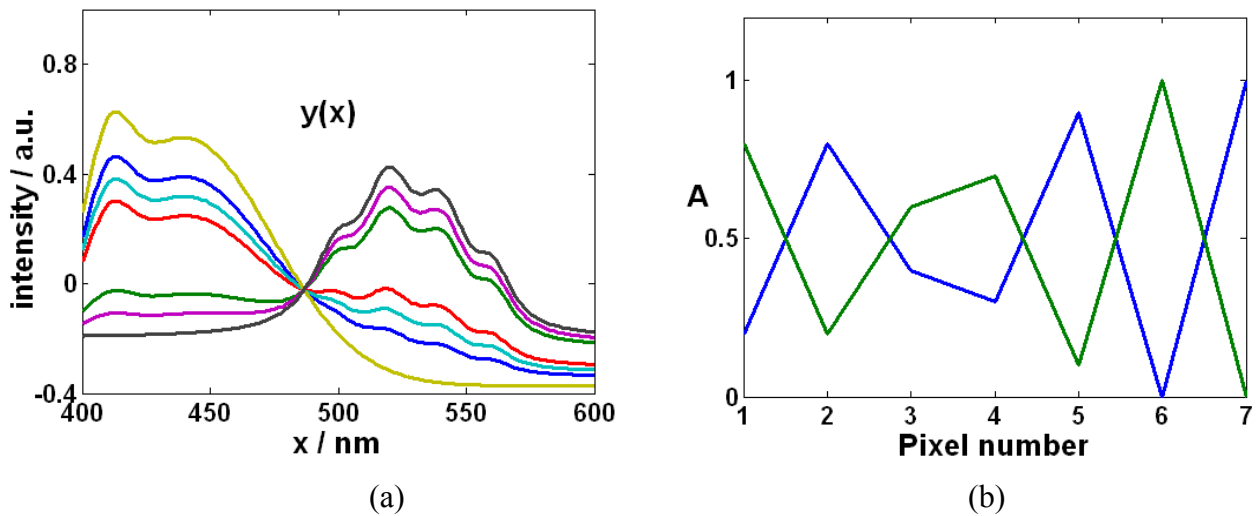

(a) (b)

**Figure 2.** The modeled spectra of the $P_1$-$P_2$ mixture and the mixing matrix A

The results of SOS analysis by using y(x) as the initial data (the spectra of the P1-P2 solution)

were given in Figure 3, where $\hat{s}_1(x)$ and $\hat{s}_2(x)$ are the pure spectra of $P_1$ and $P_2$, respectively.

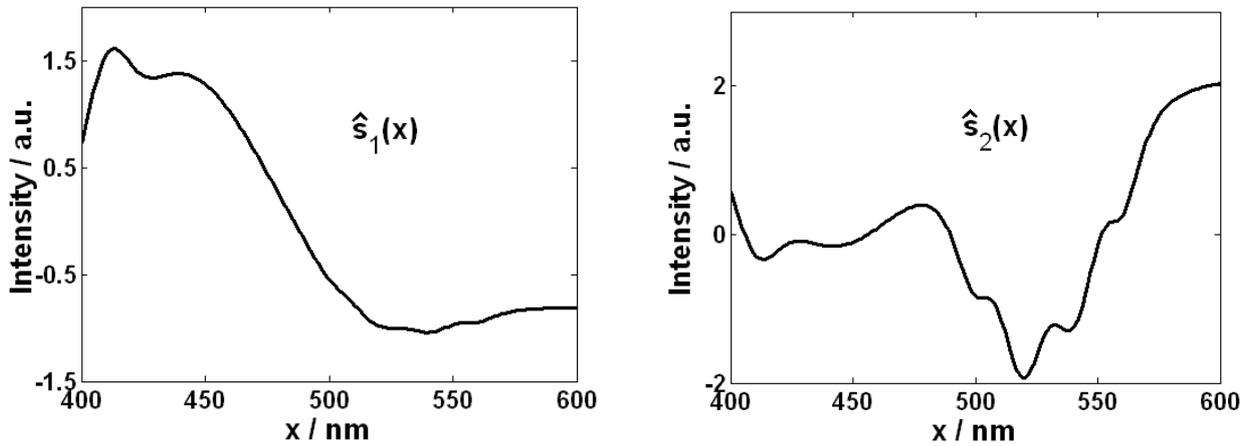

**Figure 3.** The estimated pure spectra from the initial y(x)

By comparing Figure 1 and 3, it can be seen that the patterns of the estimated pure spectra of $P_1$ and $P_2$ by SOS were quite similar to the initial ones, but the peak direction of $\hat{s}_2(x)$ was the opposite to the initial pure spectrum of $P_2$ in Figure 1.

The change profile of column vector of the mixing matrix A calculated from the estimated pure spectra by the equation (6) was shown in Figure 4, from which it can be seen that the mixing matrix A from the estimated spectra was different from the one from the initial spectra in Figure 2-b.

The difference in profile changes of the column vector between mixing matrix A from the initial spectra and that from the estimated one can be attributed to the opposite peak direction of the pure spectra.

From the above trial application of SOS in estimation of the pure spectra from the visible spectrum of their mixture, it can be concluded that one of the key problems here in SOS application for chemical mixtures is to choose the correct direction of the peaks in the estimated pure spectra of the chemical components.

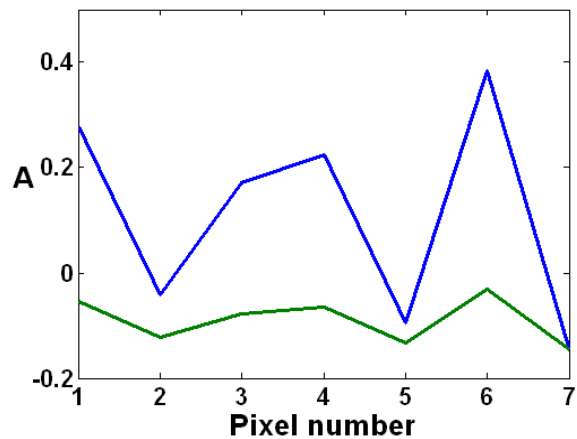

**Figure 4.** The change profile of column vector of A from the estimated spectra

To solve this problem, we propose one method for the correct peak direction as follows.

The first step of this method is the rational definition of the base line for the pure spectrum.

It is common that most of the visible spectra of chemical substances have the pattern of comparatively long and smooth line, parallel to the base line and the peaks located upwards over the line. So it is quite reasonable to choose this smooth line as the base line, where the histogram value of the spectrum is supposed to be the maximum.

Therefore, the base line $b_i$ of the $i^{th}$ pure spectrum $\hat{s}_i(x)$ is chosen as follows:

$$[\vec{z}_i, \vec{y}_i] = hist(\hat{s}_i(x))$$

$$b_i = y_{ik}$$

where k satisfies $z_{ik} = \max\{z_{ij}\}, j = 1, 2, \cdots, 10$

The second step is to find out the peaks of spectrum by interpretation of the sign of $d\hat{s}_i(x)/dx$ and to locate the maximum peak value $p_{i\max}$ and minimum peak value $p_{i\min}$.

At last, the correct peak direction is determined by using the values of $p_{i\max} - b_i$ and $p_{i\min} - b_i$ as follows:

- the correct direction

    $(\{(p_{i\max} - b_i) > 0\} \& \{(p_{i\min} - b_i) > 0\}$ or

    $\{(p_{i\max} - b_i) > 0\} \& \{(p_{i\min} - b_i) < 0\} \& \{|P_{i\max} - b_i| > |P_{i\min} - b_i|\}$

- the opposite direction

    $\{(p_{i\max} - b_i) < 0\} \& \{(p_{i\min} - b_i) < 0\}$ or

    $\{(p_{i\max} - b_i) > 0\} \& \{(p_{i\min} - b_i) < 0\} \& \{|P_{i\max} - b_i| < |P_{i\min} - b_i|\}$

In case of the opposite direction, the signs of the corresponding $\hat{s}_i(x)$ have to be changed.

**3. Results and Discussion**

The present method for choosing the correct peak direction was applied in SOS analysis for the above model and the results estimated from the initial spectra of the $P_1$-$P_2$ mixture in Figure 2 were shown in Figure 5, 6.

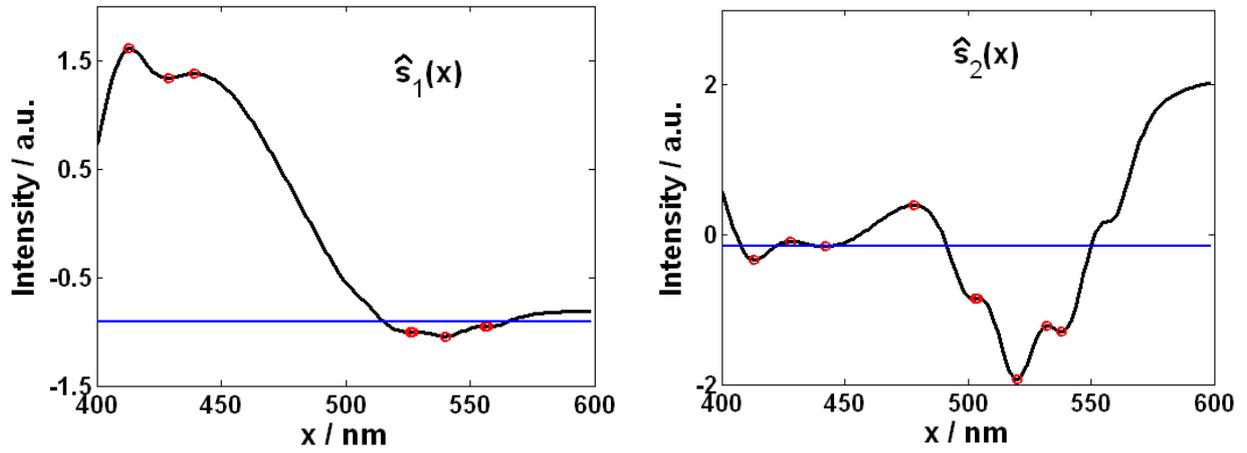

**Figure 5.** The baselines for the pure spectra of the two components

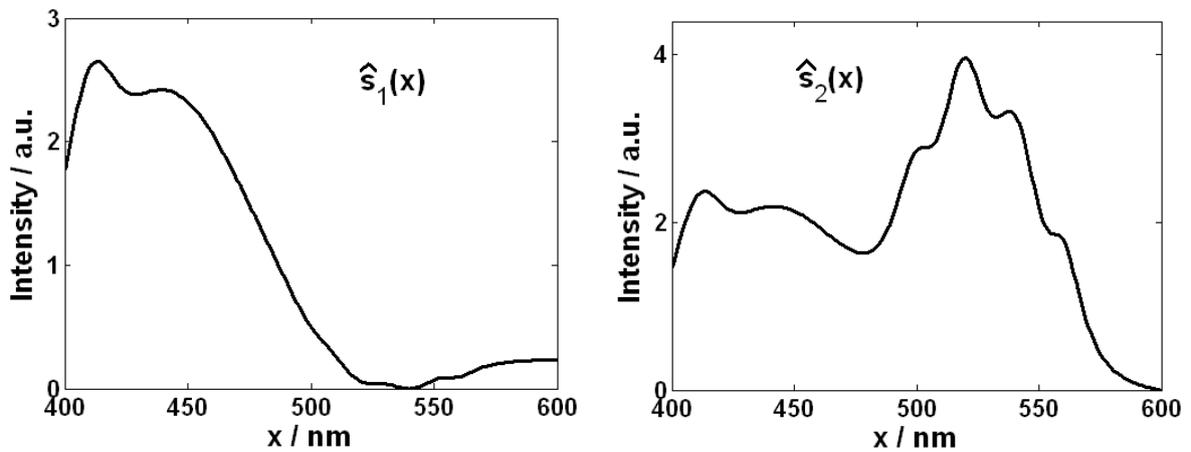

**Figure 6.** The pure spectra of $P_1$ and $P_2$ estimated by the present method

The change profile of column vector of mixing matrix A calculated from the pure spectra by the present method was shown in Figure 7.

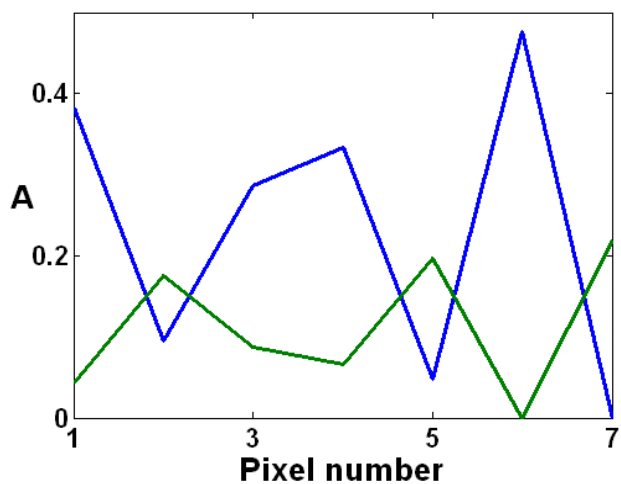

**Figure 7.** The change profile of the newly estimated column vector of A

In comparison with those in Figure 4, the change profiles in Figure 7 are in better coincidence with the initial ones in Figure 2-b.

These results showed that the proposed method for the correct peak direction in the SOS estimation of the pure spectra of the chemical components from the visible spectra of their mixture can significantly reduce errors not only in the direction of the peaks, but also in the concentration distribution analysis of chemical components in the sample mixtures.

In addition, possibility of applying the present method in multi-component systems was tested with another model solution with 3-component solutes ($P_1$, $P_2$, $P_3$), where the initial pure specrtra of $P_1$, $P_2$, and $P_3$ were modeled like those in Figure 8.

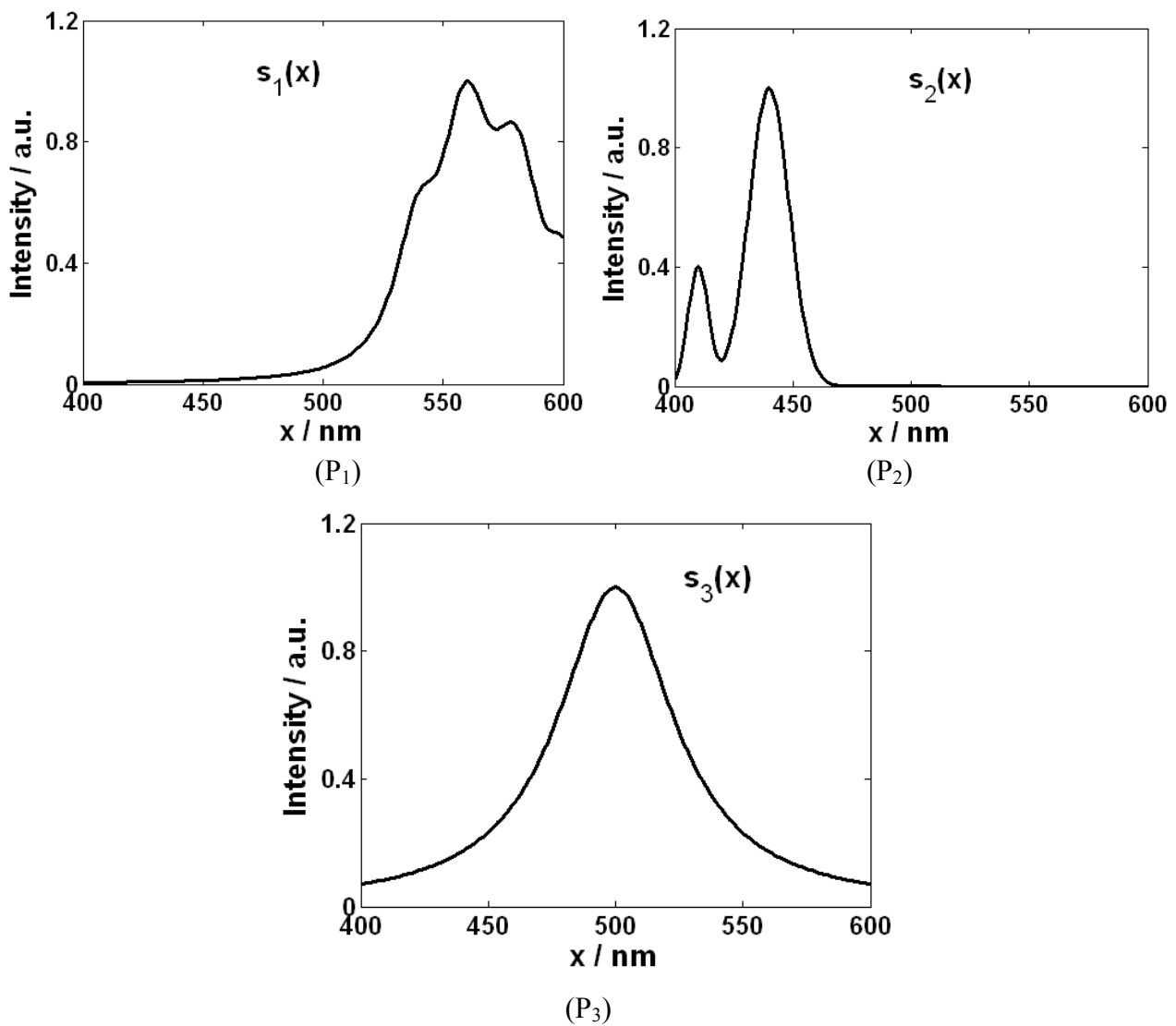

**Figure 8.** The spectral patterns of the 3 components ($P_1$, $P_2$, $P_3$)

Their mixing matrix A was assumed as follows:

A = [ 0.2   0.7   0.1 ; 0.6   0.1   0.3; 0.2   0.4   0.4 ;   0.1   0.5   0.4;
      0.7   0.1   0.2; 0.0   0.7   0.3;   0.7   0.2   0.1 ]

The spectra of their mixture ($P_1$-$P_2$-$P_3$) was modeled by $y(x) = As(x)$, $s(x) = (s_1(x), s_2(x), s_3(x))^T$ (Figure 9).

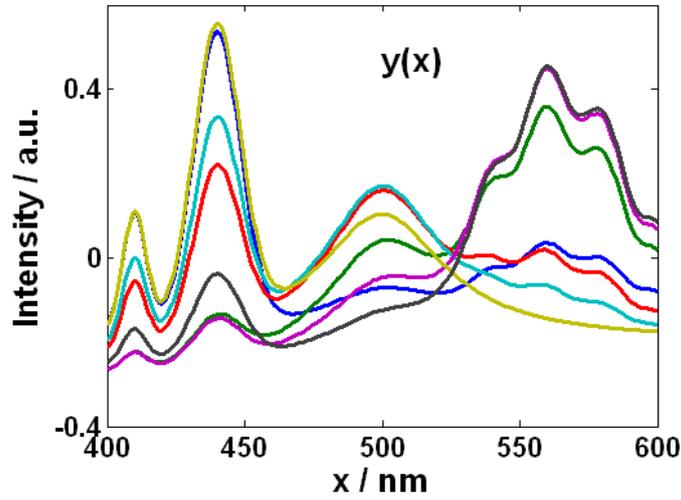

**Figure 9.** The simulated spectra of the $P_1$-$P_2$-$P_3$ mixture

The estimated pure spectra for the $P_1$-$P_2$-$P_3$ mixture by the present method were given in Figure 10.

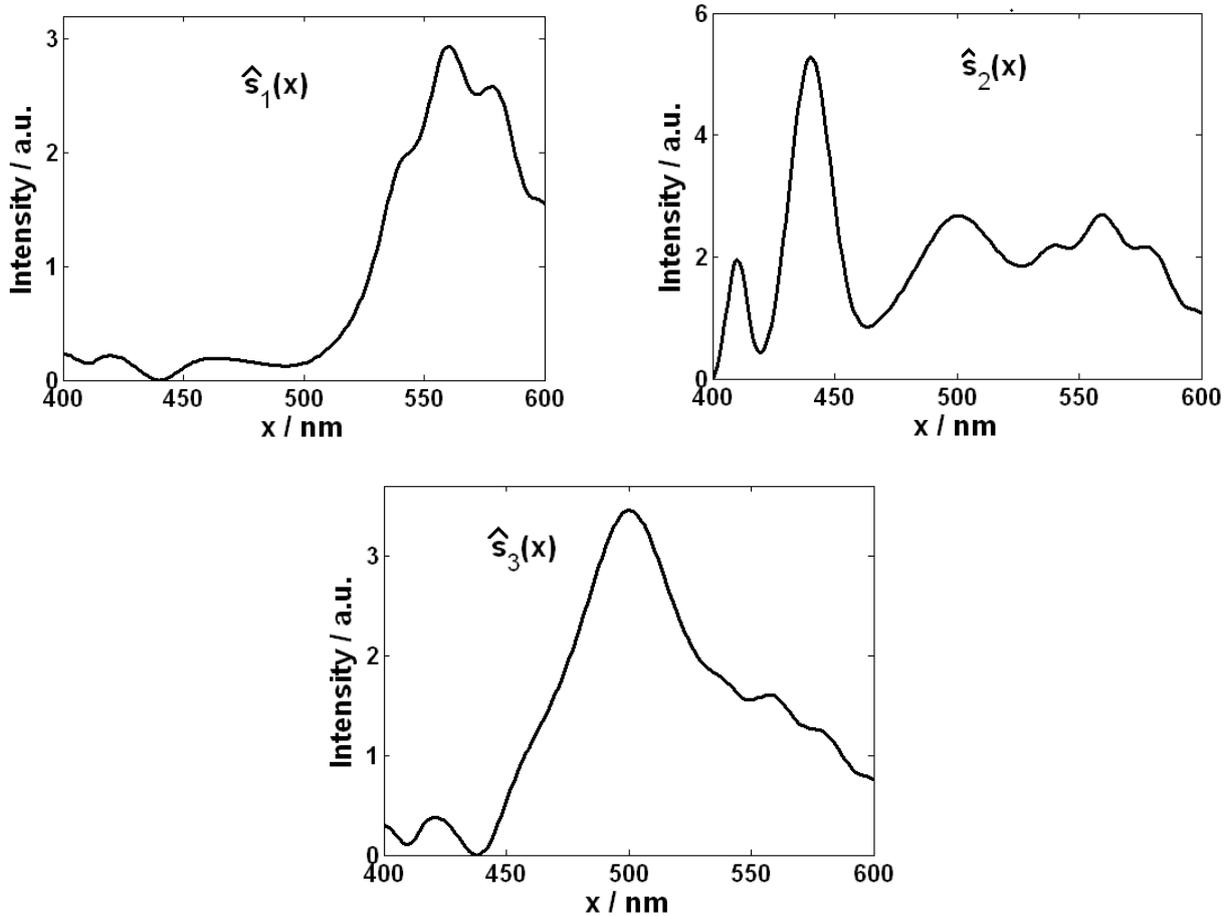

**Figure 10.** The estimated pure spectra of $P_1$-$P_2$-$P_3$ mixture

These results imply that the present method can offer the wider application of SOS analysis in the investigation of behavior of the individual components in the multi-component chemical systems.

## 5. Conclusion

Second-Order Statistics seems to be one of the prosperous spectral image analysis methods in the investigation of the complex chemical systems. Trial application of SOS in estimation of the pure spectra of the individual chemical components from the visible spectra of one model mixture composed of two components showed that it may give the unreasonable pure spectra of the chemical components with the opposite peak direction, which can be a reason of increasing the errors in determination of the true pure spectra of the components. Here in this paper presented was one method for judgment of the correct direction of peaks in the pure spectra of the component obtained by SOS analysis of the spectra of their mixture, where the base line defined from the histogram and the maximum- and minimum-values of the peaks were used for choosing the reasonable peak direction. The present method was applied in SOS analysis of the pure spectra of the components from the visible spectra of the two- and three-component chemical mixtures and gave the better results in the peak direction and concentration distribution, compared with the common SOS analysis. More detailed researches for SOS application in the spectral image analysis of the complicated multi-component chemical systems seems to have good prospect in practice of chemistry and SOS itself.

## References

[1] Antoine Souloumiac, "A Stable and Efficient Algorithm for Difficult Non-orthogonal Joint Diagonalization Problems", *19th European Signal Processing Conference (EUSIPCO 2011) Barcelona, Spain,* 2011, August 29 - September 2.

[2] Adel Belouchrani and Andrzej Cichocki, "A Robust Whitening Procedure in Blind Source Separation Context", *Electronics Letters,* 2000, 36, 2050-2051.

[3] A. Ikhlef, R.; Iferroujene, A.; Boudjellal, K.; Abed-Meraim, A.; Adel Belouchrani, "Constant Modulus Algorithms Using Hyperbolic Givens Rotation", 2013, *arXiv*: 1306.4128vl.

[4] Mesloub Ammar; Karim Abed-Meraim; Adel Belouchrani, "A New Algorithm for Complex Non orthogonal Joint Diagonalization Based on Shear and Givens Rotations", 2013, *arXiv*: 1306.0331vl.

[5] Lucas Parra; Paul Sajda, "Blind Source Separation via Generalized Eigenvalue Decomposition", *Journal of Machine Learning Research,* 2003, 4, 1261- 1269.

[6] Massart, D.L.; Vandeginste, B.G.M.; Buydens, L.M.C.; Jong, S.; Lewi, P.J.; Smeyers-Verbeke, J.,

"Handbook of Chemometrics and Qualimetrics, in Data Handling in Science and Technology", Elsevier, 1997, 20.

[7] Keshava, N.; Mustard, J. F., "Spectral Unmixing", *IEEE Signal Proc. Mag.* 19, 2002, 44–57.

[8] Reiner Salzer and Heinz W. Siesler, "Infrared and Raman Spectroscopic Imaging", WILEY-VCH, 2009, 90-101.

[9] Klaus-Robert Muller, "Blind Source Separation Techniques for Decomposing Event-Related Brain Signals", *International Journal of Bifurcation and Chaos,* 2004, 2, 773-791.

[10] Lucas Parra and Clay Spence, "Convolutive Blind Separation on Non-Stationary Sources", *IEEE Transactions on Speech and Audio Processing*, 2000, 8, 3, 320-327.

[11] R. R. Gharieb and A. Cichocki, "Second-Order Statistics Based Blind Source Separation Using a Bank of Subband Filters", *Digital Signal Processing,* 2003, 13, 252-274.

[12] Anne Ferreol; Pascal Chevalier; Laurent Albera, "Second-Order Blind Separation of First- and Second-Order Cyclostationary Sources-Application to AM, FSK, CPFSK, and Deterministic Sources", *IEEE Transactions on Signal Processing*, 2004, 52, 4, 845-861.

[13] Herbert Buchner; Robert Aichner; Walter Kellermann, "A Generalization of Blind Source Separation Algorithms for Convolutive Mixtures Based on Second-Order Statistics", *IEEE Transactions on Speech and Audio Processing*, 2005, 13, 1, 120-134.

[14] Adel Belouchrani; Karim Abed-Meraim; Jean-Francois Cardoso; Eric Moulines, "A Blind Source Separation Technique Using Second-Order Statistics", *IEEE Transactions on Signal Processing*, 1997, 45, 2, 434-444.

[15] Pierre Jallon and Antoine Chevreuil, "Separation of Instantaneous Mixtures of Cyclostationary Sources with Application to Digital Communication Signals", *14th European Signal Processing Conference (EUSIPCO 2006)*, Florence, Italy, 2006, September 4-8.

[16] Foad Ghaderi; Hamid R. Mohseni; Saeid Sanei, "A Fast Second Order Blind Identification Method for Separation of Periodic Sources", *18th European Signal Processing Conference (EUSIPCO 2010), Aalborg, Denmark,* 2010, August 23-27.

[17] Antoni, J. S.; Chauhan, "Second Order Blind Source Separation Techniques (SO-BSS) and Their Relation to Stochastic Subspace Identification (SSI) Algorithm", 2010, *Proceedings of the IMAC-XXVIII*, February 1-4

[18] MATLAB, Version 7.0; MathWorks, Inc., 2004.